\documentclass{article}

\usepackage{microtype}
\usepackage{graphicx}
\usepackage{subcaption}
\usepackage{booktabs} 

\usepackage{hyperref}

\usepackage[accepted]{icml2026}

\usepackage{amsmath}
\usepackage{amssymb}
\usepackage{mathtools}
\usepackage{amsthm}
\usepackage{times}
\usepackage{latexsym}
\usepackage{inconsolata}
\usepackage{adjustbox}
\usepackage{graphicx}
\usepackage[table]{xcolor}
\usepackage{booktabs}
\usepackage{array}
\usepackage{fontawesome5}
\usepackage{subcaption}
\usepackage{multirow}
\usepackage{adjustbox}
\usepackage{amssymb}
\usepackage{float}
\usepackage{caption}
\usepackage[utf8]{inputenc} 
\usepackage[T1]{fontenc}    
\usepackage{hyperref}       
\usepackage{url}            
\usepackage{xurl}
\usepackage{booktabs}       
\usepackage{amsfonts}       
\usepackage{nicefrac}       
\usepackage{microtype}      
\usepackage{xcolor}         
\usepackage{listings}

\usepackage{fancyhdr}

\newcommand{\cmark}{\textcolor{green}{\checkmark}}
\newcommand{\xmark}{\textcolor{red}{\texttimes}}

\usepackage[capitalize,noabbrev]{cleveref}

\theoremstyle{plain}

\theoremstyle{definition}

\theoremstyle{remark}

\usepackage[disable,textsize=tiny]{todonotes}

\icmltitlerunning{AMVICC: A Novel Benchmark for Cross-Modal Failure Mode Profiling for VLMs and IGMs}

\begin{document}

\twocolumn[
  \icmltitle{AMVICC: A Novel Benchmark for \\
    Cross-Modal Failure Mode Profiling for VLMs and IGMs}



  \icmlsetsymbol{equal}{*}

  \begin{icmlauthorlist}
    \icmlauthor{Aahana Basappa}{equal,centen,ai}
    \icmlauthor{Pranay Goel}{equal,lt,ai}
    \icmlauthor{Anusri Karra}{hsn,ai}
    \icmlauthor{Anish Karra}{hsn,ai}
    \icmlauthor{Asa Gilmore}{ai}
    \icmlauthor{Kevin Zhu}{ai}
  \end{icmlauthorlist}

  \icmlaffiliation{centen}{Centennial High School, Frisco, Texas, USA}
  \icmlaffiliation{lt}{Lebanon Trail High School, Frisco, Texas, USA}
  \icmlaffiliation{hsn}{West Windsor-Plainsboro High School, Princeton Junction, New Jersey, USA}
  \icmlaffiliation{ai}{Algoverse AI Research, Palo Alto, California, USA}

  \icmlcorrespondingauthor{Aahana Basappa}{aahana.basappa@gmail.com}
  \icmlcorrespondingauthor{Pranay Goel}{pg2037393@gmail.com}

  \icmlkeywords{Machine Learning, ICML, Computer Vision and Pattern Recognition, Vision Language Models, Image Generation Models, Visual Reasoning, Multimodal Systems, Cross-Modal Evaluation, Failure Mode Analysis}

  \vskip 0.3in
]



\printAffiliationsAndNotice{\icmlEqualContribution}

\begin{abstract}
We investigate visual reasoning limitations of both multimodal large language models (MLLMs) and image generation models (IGMs) by creating a novel benchmark to systematically compare failure modes across image-to-text and text-to-image tasks, enabling cross-modal evaluation of visual understanding. Despite rapid growth in machine learning, vision language models (VLMs) still fail to understand basic visual concepts such as object orientation, quantity, and spatial relationships, which highlights gaps in elementary visual reasoning. By adapting MMVP benchmark questions into explicit and implicit prompts, we create \textit{AMVICC}, a novel benchmark for profiling failure modes across various modalities. After testing 11 MLLMs and 3 IGMs in 9 categories of visual reasoning, our results show that failure modes are often shared between models and modalities. However, certain failures are model-specific and modality-specific, and this can potentially be attributed to various factors. IGMs consistently struggle to manipulate specific visual components in response to prompts, especially in explicit prompts, suggesting poor control over fine-grained visual attributes. Our findings apply most directly to the evaluation of existing state-of-the-art models on structured visual reasoning tasks. This work lays the foundation for future cross-modal alignment studies, offering a framework to probe whether image generation and visual interpretation failures stem from shared limitations. These insights can guide future improvements in unified vision-language modeling.
\end{abstract}

\section{Introduction}
Recently, multimodal large language models have improved significantly and have shown proficiency in several fields with emergent capabilities \citep{stabled}. However, recent work has highlighted that, despite their strengths in visual reasoning, instruction following, and image understanding, many MLLMs fail to consistently and accurately answer straightforward visual understanding questions that most humans find trivial \citep{vlmlimitations}. The extensive visual shortcomings of MLLMs and VLMs have been defined and tested in benchmarks such as MediConfusion, GMAI-MMBench, and MMVP \citep{mediconfusion, gmai, mmvp}.

Compared to other generative model modalities, IGMs are steadily improving: Google's Gemini 2.5 Flash Image and OpenAI’s DALL·E 3 revolutionize instruction-following and realism within image generation \citep{gemini25flash, dalle3}. However, despite their drastic growth, IGMs demonstrate similar elementary failures in generating images that align with given prompts, especially those with a complex combination of entities, attributes, and spatial relationships \citep{igmfailures, spatial}. Several benchmarks, such as VisuLogic, VISOR, and T2I-CompBench++ \citep{visulogic, visor, compbench}, have attempted to classify failure modes to identify prospective points of improvement. These benchmarks evaluate failure modes across categories consisting of quantitative shifts, attribute comparisons, and spatial relationships.

\begin{figure*}
    \centering
    \includegraphics[width=320px, height=151px]{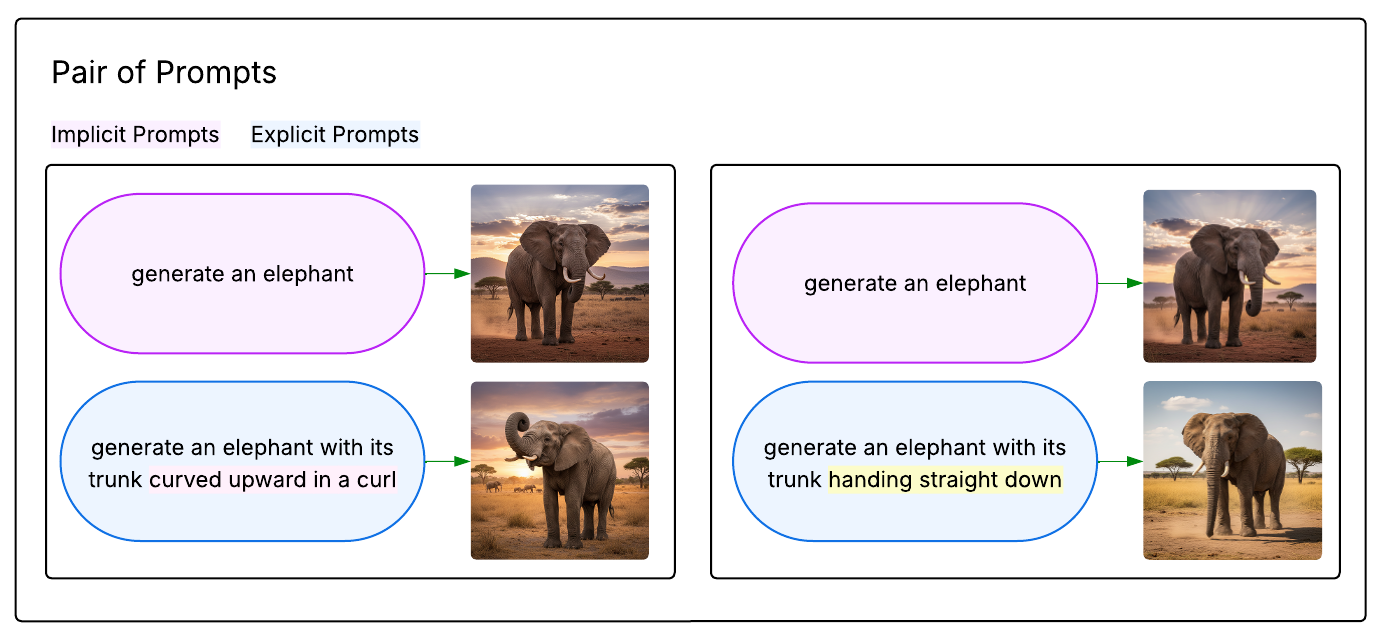}
    \vskip 0.1in
    \caption{Comparison of implicit prompts to explicit prompts for the 2 images in a pair.}
    \label{fig:placeholder}
\end{figure*}

However, there is a notable lack of research comparing visual reasoning and image generation between MLLMs and IGMs, respectively. In this paper, we extend the work done by \textit{Eyes Wide Shut? Exploring the Visual Shortcomings of Multimodal LLMs} to profile the cross-modal failure modes in visual reasoning and recognition of MLLMs and IGMs \citep{mmvp}. \textit{Eyes Wide Shut? Exploring the Visual Shortcomings of Multimodal LLMs} presents a benchmark, MultiModal Visual Patterns (MMVP), to evaluate the elementary visual shortcomings of MLLMs through a series of Yes or No questions on spatial understanding, textual context, perspective, and presence of features, among others. A model is graded on its ability to correctly identify the difference between 2 images with similar CLIP values but obvious variations in content. Based on accuracy, the authors are able to conclude that MLLMs struggle with apparent straightforward disparities across several different categories of comprehension. We create matching image generation prompts based on the MMVP benchmark to evaluate failure mode similarities between the two modalities. Through these tests, we hope to uncover insights into the elementary visual shortcomings of MLLMs and IGMs. In this paper, we introduce a novel benchmark, \textbf{A}ssessment of \textbf{M}odality-Specific \textbf{V}isual \textbf{I}ntelligence \textbf{C}omprehension and \textbf{C}reation (AMVICC), to evaluate the failure modes of multimodal large language models and image generation models with the same contextual input and provide analysis of tests completed on current state-of-the-art models.

\section{Methods}
\label{methods}
In this section, we explain our evaluation of the following Vision Language Models: Meta: Llama 3.2 90B Vision Instruct (90 billion parameters), Meta: Llama 4 Maverick (17 billion active parameters and 128 experts), Meta: Llama 4 Scout (17 billion active parameters and 16 experts), xAI: Grok 4, Google: Gemma 3 27B (27 billion parameters), Google: Gemini 2.5 Pro, OpenAI: GPT-4o, Qwen: Qwen2.5 VL 72B Instruct (72 billion parameters), Mistral: Pixtral Large 2411 (124 billion parameters), Anthropic: Claude Opus 4.1, and Anthropic: Claude Sonnet 4, all based on a modified version of the MMVP benchmark \citep{llama32, llama4, grok4, gemma3, gemini2.5, gpt4o, qwen, pixtral, opus, sonnet}. To our modified version, we assign categories to MMVP benchmark questions to match the tasks of the visual understanding questions. Along with the VLMs, we also evaluate the following Image Generation Models: OpenAI: DALL·E 3, Google: Gemini 2.5 Flash Image, and Stability AI: Stable Diffusion 3.5 Large (8.1 billion parameters), all based on \textit{AMVICC}, which contains prompts constructed to mirror the questions from MMVP with corresponding categories \citep{dalle3, gemini25flash, stabled}. VLMs are chosen to provide a variance across open-source and closed-source models while also providing variability across model size, architecture, and training methods. Due to limited access and availability, there is a smaller selection of state-of-the-art IGMs; we are only able to choose 3 models with variance across providers, training data, architecture, and size.

\subsection{Prompting Procedure} \label{prompting_procedure}
To evaluate model performance in both directions (image → text and text → image), we use 300 original MMVP benchmark questions, and we create 600 additional prompts (found in Appendix \ref{appendix_a}) to probe specific failure modes.

\begin{itemize}
   \item \textbf{For Vision Language Models:} For VLMs, the benchmark questions are paired with MMVP images, and the resulting answers are graded by OpenAI's GPT-4 \citep{gpt-tech} to determine model accuracy.
   \item \textbf{For Image Generation Models:} For image generation, we (four of the authors) design hand-crafted explicit and implicit prompts derived from the MMVP questions. This is done in order to test corresponding tasks in image generation models with 2 consequent checks for correct structure and prompting style.
\end{itemize}
These mixed evaluation methods are utilized because VLMs, while proven to be accurate with text, are known to be inaccurate when evaluating images for positioning and elementary understanding. This methodology also mirrors that of MMVP, which summarizes outputs from the VLMs into multiple-choice answers (e.g., (a) or (b)). Our implicit prompts are created by defining the generalized situation between a pair of images in order to establish the foundation of a model's ability to generate the background. Afterwards, each explicit prompt, correlating to an MMVP question, adds the element required by the correct answer choice of the corresponding MMVP question. Explicit prompts clearly define the required visual concept, while implicit prompts use more natural, generalized phrasing to create a prompt relevant to both the question and the correct answer choice. There are a total of 600 prompts, consisting of 300 implicit prompts and 300 explicit prompts. Each implicit prompt tests an image generation model’s ability to generate a scenario, while each explicit prompt tests its ability to change the generated image to satisfy a specific newly-added component (e.g., \textit{``dog in grass''} is implicit, whereas \textit{``dog in grass looking to the right''} is explicit). Our creation of explicit prompts is similar to the way MMVP tests a model's ability to visually understand a specific component.

\begin{figure}[H]
    \centering
    \includegraphics[width=170px, height=228px]{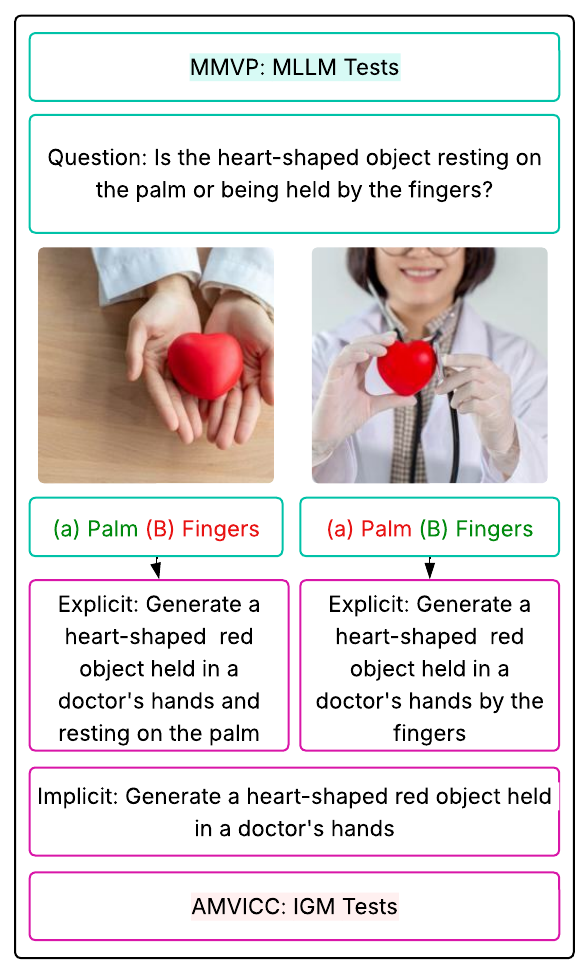}
    \vskip 0.1in
    \caption{Diagram of the AMVICC Creation Pipeline: We create implicit prompts based on the general scenario introduced by the question and create explicit prompts by adding specifics in line with the specific answer choice for that image ID.}
    \label{fig:placeholder2}
\end{figure}

\subsection{Evaluation}

To systematically evaluate the performance of both vision language models and image generation models, we determine the success and failure of each question-answering task. We design a rubric that defines success and failure based on the intended visual understanding goals of MMVP. We begin by testing various VLMs (detailed in Section \ref{methods}) based on the 300 questions that are present in the original MMVP dataset. We then move on to testing various IGMs (detailed in Section \ref{methods}) with the additional prompts that we create based on those questions. By performing this, we are able to highlight certain aspects that automated benchmarks might not have been able to catch. This is intended to measure the visual understanding goals of MMVP and of each prompt based on the visual understanding of \textit{AMVICC}.

Images generated by IGMs are evaluated differently depending on whether the prompt is implicit or explicit. An implicit image is considered correct if it satisfies all components of the provided prompt, regardless of whether the image matches the corresponding question's distinction. An explicit image is considered correct if it contains the specific feature that the prompt asks for based on the corresponding visual understanding question and category from our modified MMVP benchmark. Each image is scored by human evaluators and double-checked for accuracy in order to mitigate bias and ensure correct grading.

\section{Results}
\label{results}

We evaluate the accuracy of 11 multimodal LLMs in visual understanding and reasoning tasks via the MMVP benchmark \citep{mmvp}. VLM accuracy scores depict the models' proficiency across questions of the 9 visual reasoning categories. After our evaluation of these 11 models, we extend our experiments to 3 image generation models using our AMVICC benchmark to evaluate each IGM's proficiency in generating images across the 9 categories. Failure modes are defined as individual accuracies below 80\% and pair accuracies below 70\%. This applies to both MLLMs and IGMs. Each pair of questions is only considered correct if both questions are answered correctly or both images generated are aligned with their respective explicit prompts.

\subsection{MLLM Score Analysis} \label{mllm_score_analysis}
Many of the models share the same failure modes; however, some of the models have failure modes that served as outliers. For example, in both the Orientation and Direction and the Quantity and Count categories, the individual VLM accuracies for xAI: Grok 4 are 40.00\% and 50.00\%, respectively (see Table \ref{tab:vit_performance}). In the Viewpoint and Perspective category, xAI: Grok 4 and Anthropic: Claude Sonnet 4 are outliers, both attaining an accuracy of 55.56\%, a notable 16.66\% difference from the next highest accuracy. That being said, an opposite trend is evident in Table \ref{tab:vit_performance2}, which displays the pair VLM accuracies. Instead of the outliers constituting failure modes, they are the highest accuracies for models such as Meta: Llama 3.2 90B Vision Instruct and Meta: Llama 4 Maverick. This is exhibited in the Positional and Relational Context as well as the Viewpoint and Perspective categories for Meta: Llama 3.2 90B Vision Instruct. This trend is apparent in the Quantity and Count category for Meta: Llama 4 Maverick. Consequently, this trend highlights the fact that certain models succeed where either all or most of the other models fail. Furthermore, most MLLMs fail in similar contexts, particularly in Positional and Relational Context and Quantity and Count. Additional common failure modes include State and Condition, Orientation and Direction, and Viewpoint and Perspective. However, model-specific failure modes occur as well, with only xAI: Grok 4 failing on Color and Appearance, and four models out of eleven (Google: Gemini 2.5 Pro, xAI: Grok 4, Google: Gemma 3 27B, and Anthropic: Claude Opus 4.1) failing on visual reasoning within the category of Structural and Physical Characteristics (see Table \ref{tab:vit_performance}). This suggests a variance of failure modes for certain models in addition to the common failure modes.

\begin{figure}
    \centering
    \includegraphics[width=\columnwidth]{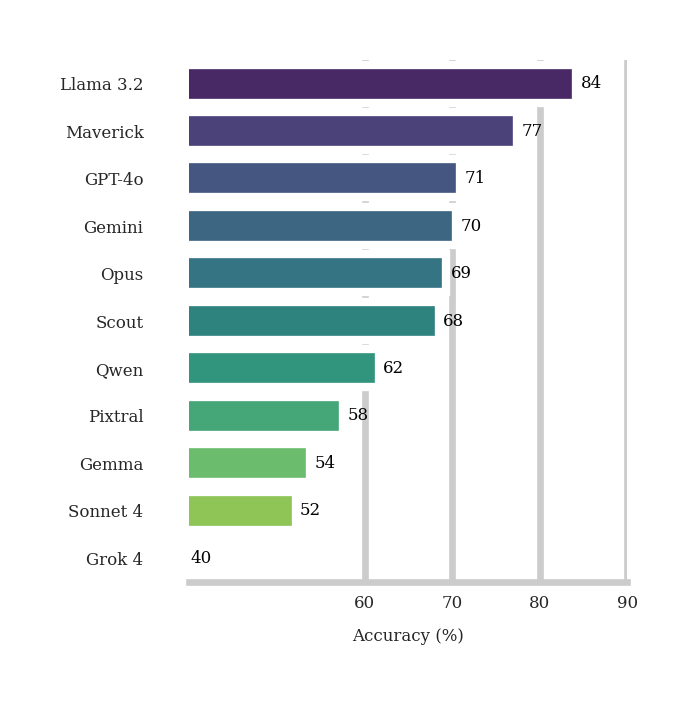}
    \caption{Benchmark results of MLLMs: We evaluate pair accuracy across 11 models based on the questions and corresponding images from the MMVP dataset.}
    \label{fig:placeholder3}
\end{figure}

\begin{table*}[htbp]
\caption{Individual VLM Accuracies: Based on images and associated questions from the MMVP dataset. \colorbox{gray!30}{Failure modes} are highlighted where present, based on definitions (see Section \ref{results}). \colorbox{cyan!25}{Highest non-failure mode accuracies in each category} are highlighted, largely among a subset of models. We use symbols as a representation for all 9 categories: \faSync: State and Condition, \faCog: Structural and Physical Characteristics, \faCompass: Orientation and Direction, \textbf{A}: Text, \faSort: Quantity and Count, \faMapPin: Positional and Relational Context, \faSearch: Presence of Specific Features, \faCamera: Viewpoint and Perspective, and \faPalette: Color and Appearance. Based on the number of failure modes, it's evident that Quantity and Count, as well as Positional and Relational Context are the two categories that the VLMs struggle the most with.}
\vskip 0.1in
\centering
\scriptsize
\setlength{\tabcolsep}{5pt}
\begin{tabular}{l|c|c|c|c|c|c|c|c|c|c|c}
\toprule
\textbf{Model} & \textbf{Params} & \faSync & \faCog & \faCompass & \textbf{A} & \faSort & \faMapPin & \faSearch & \faCamera & \faPalette & \textbf{Model} \\
& \textbf{Size (B)} & & & & & & & & & & \textbf{Average} \\
\midrule
OpenAI: GPT-4o \citep{gpt4o} & $-$ & \cellcolor{gray!30}77.78 & 83.33 & 83.33 & 85.71 & \cellcolor{gray!30}75.00 & \cellcolor{gray!30}78.13 & 91.43 & 94.44 & \cellcolor{cyan!25}96.43 & \textbf{85.06} \\
Google: Gemini 2.5 Pro \citep{gemini2.5} & $-$ & \cellcolor{gray!30}79.63 & \cellcolor{gray!30}76.67 & 86.67 & \cellcolor{cyan!25}92.86 & \cellcolor{gray!30}79.17 & \cellcolor{gray!30}78.13 & 88.57 & 88.89 & 89.29 & \textbf{84.43} \\
Qwen: Qwen2.5 VL 72B Instruct \citep{qwen} & $72$ & \cellcolor{gray!30}74.07 & 83.33 & \cellcolor{gray!30}73.33 & 85.71 & \cellcolor{gray!30}79.17 & \cellcolor{gray!30}71.88 & 90.00 & \cellcolor{gray!30}72.22 & 82.14 & \textbf{79.09} \\
Mistral: Pixtral Large 2411 \citep{pixtral} & $124$ & 83.33 & 86.67 & \cellcolor{gray!30}66.67 & \cellcolor{gray!30}71.43 & \cellcolor{gray!30}79.17 & \cellcolor{gray!30}71.88 & 88.57 & \cellcolor{gray!30}72.22 & 85.71 & \textbf{78.41} \\
xAI: Grok 4 \citep{grok4} & $-$ & \cellcolor{gray!30}62.96 & \cellcolor{gray!30}73.33 & \cellcolor{gray!30}40.00 & \cellcolor{gray!30}64.29 & \cellcolor{gray!30}50.00 & \cellcolor{gray!30}50.00 & 82.86 & \cellcolor{gray!30}55.56 & \cellcolor{gray!30}67.86 & \textbf{60.76} \\
Google: Gemma 3 27B \citep{gemma3} & $27$ & \cellcolor{gray!30}68.52 & \cellcolor{gray!30}73.33 & \cellcolor{gray!30}66.67 & \cellcolor{gray!30}78.57 & \cellcolor{gray!30}70.83 & \cellcolor{gray!30}68.75 & 90.00 & \cellcolor{gray!30}72.22 & 89.29 & \textbf{75.35} \\
Meta: Llama 3.2 90B Vision Instruct \citep{llama32} & $90$ & 87.04 & \cellcolor{cyan!25}96.67 & \cellcolor{cyan!25}90.00 & 85.71 & 83.33 & \cellcolor{cyan!25}90.63 & \cellcolor{cyan!25}97.14 & \cellcolor{cyan!25}100.00 & \cellcolor{cyan!25}96.43 & \textbf{91.88} \\
Meta: Llama 4 Maverick \citep{llama4} & \texttt{17Bx128E} & \cellcolor{cyan!25}88.89 & 93.33 & 86.67 & \cellcolor{cyan!25}92.86 & \cellcolor{cyan!25}95.83 & \cellcolor{gray!30}71.88 & 90.00 & \cellcolor{gray!30}77.78 & 89.29 & \textbf{87.39} \\
Meta: Llama 4 Scout \citep{llama4} & \texttt{17Bx16E} & 81.48 & 93.33 & \cellcolor{gray!30}70.00 & 85.71 & \cellcolor{gray!30}79.17 & \cellcolor{gray!30}75.00 & 95.71 & \cellcolor{gray!30}72.22 & 92.86 & \textbf{82.83} \\
Anthropic: Claude Opus 4.1 \citep{opus} & $-$ & 83.33 & \cellcolor{gray!30}76.67 & 83.33 & 85.71 & \cellcolor{gray!30}75.00 & 81.25 & 87.14 & 94.44 & 85.71 & \textbf{83.62} \\
Anthropic: Claude Sonnet 4 \citep{sonnet} & $-$ & \cellcolor{gray!30}77.78 & 80.00 & \cellcolor{gray!30}60.00 & \cellcolor{gray!30}78.57 & \cellcolor{gray!30}70.83 & \cellcolor{gray!30}75.00 & 87.14 & \cellcolor{gray!30}55.56 & 89.29 & \textbf{74.91} \\
\midrule
\textbf{Category Average} & & \textbf{78.62} & \textbf{83.33} & \textbf{73.97} & \textbf{82.40} & \textbf{75.23} & \textbf{74.78} & \textbf{89.87} & \textbf{77.78} & \textbf{87.66} & \textbf{80.34} \\
\bottomrule
\end{tabular}
\label{tab:vit_performance}
\end{table*}

\begin{table*}[htbp]
\caption{Pair VLM Accuracies: Based on images and associated questions from the MMVP dataset. \colorbox{gray!30}{Failure modes} are highlighted where present, based on definitions (see Section \ref{results}). \colorbox{cyan!25}{Highest non-failure mode accuracies in each category} are highlighted, largely among a subset of models.}
\vskip 0.1in
\centering
\scriptsize
\setlength{\tabcolsep}{5pt}
\begin{tabular}{l|c|c|c|c|c|c|c|c|c|c|c}
\toprule
\textbf{Model} & \textbf{Params} & \faSync & \faCog & \faCompass & \textbf{A} & \faSort & \faMapPin & \faSearch & \faCamera & \faPalette & \textbf{Model} \\
& \textbf{Size (B)} & & & & & & & & & & \textbf{Average} \\
\midrule
OpenAI: GPT-4o \citep{gpt4o} & $-$ & \cellcolor{gray!30}62.96 & \cellcolor{gray!30}66.67 & \cellcolor{gray!30}66.67 & 71.43 & \cellcolor{gray!30}50.00 & \cellcolor{gray!30}56.25 & 82.86 & 88.89 & \cellcolor{cyan!25}92.86 & \textbf{70.95} \\
Google: Gemini 2.5 Pro \citep{gemini2.5} & $-$ & \cellcolor{gray!30}66.67 & \cellcolor{gray!30}60.00 & 73.33 & \cellcolor{cyan!25}85.71 & \cellcolor{gray!30}58.33 & \cellcolor{gray!30}56.25 & 77.14 & 77.78 & 78.57 & \textbf{70.42} \\
Qwen: Qwen2.5 VL 72B Instruct \citep{qwen} & $72$ & \cellcolor{gray!30}59.26 & 73.33 & \cellcolor{gray!30}53.33 & 71.43 & \cellcolor{gray!30}58.33 & \cellcolor{gray!30}50.00 & 80.00 & \cellcolor{gray!30}44.44 & \cellcolor{gray!30}64.29 & \textbf{61.60} \\
Mistral: Pixtral Large 2411 \citep{pixtral} & $124$ & \cellcolor{gray!30}66.67 & 73.33 & \cellcolor{gray!30}40.00 & \cellcolor{gray!30}42.86 & \cellcolor{gray!30}58.33 & \cellcolor{gray!30}43.75 & 77.14 & \cellcolor{gray!30}44.44 & 71.43 & \textbf{57.55} \\
xAI: Grok 4 \citep{grok4} & $-$ & \cellcolor{gray!30}37.04 & \cellcolor{gray!30}53.33 & \cellcolor{gray!30}33.33 & \cellcolor{gray!30}42.86 & \cellcolor{gray!30}25.00 & \cellcolor{gray!30}25.00 & 71.43 & \cellcolor{gray!30}33.33 & \cellcolor{gray!30}35.71 & \textbf{39.67} \\
Google: Gemma 3 27B \citep{gemma3} & $27$ & \cellcolor{gray!30}44.44 & \cellcolor{gray!30}46.67 & \cellcolor{gray!30}33.33 & 71.43 & \cellcolor{gray!30}41.67 & \cellcolor{gray!30}43.75 & 80.00 & \cellcolor{gray!30}44.44 & 78.57 & \textbf{53.81} \\
Meta: Llama 3.2 90B Vision Instruct \citep{llama32} & $90$ & \cellcolor{cyan!25}77.78 & \cellcolor{cyan!25}93.33 & \cellcolor{cyan!25}80.00 & 71.43 & \cellcolor{gray!30}66.67 & \cellcolor{cyan!25}81.25 & \cellcolor{cyan!25}94.29 & \cellcolor{cyan!25}100.00 & \cellcolor{cyan!25}92.86 & \textbf{84.18} \\
Meta: Llama 4 Maverick \citep{llama4} & \texttt{17Bx128E} & \cellcolor{cyan!25}77.78 & 86.67 & 73.33 & \cellcolor{cyan!25}85.71 & \cellcolor{cyan!25}91.67 & \cellcolor{gray!30}56.25 & 80.00 & \cellcolor{gray!30}66.67 & 78.57 & \textbf{77.41} \\
Meta: Llama 4 Scout \citep{llama4} & \texttt{17Bx16E} & \cellcolor{gray!30}66.67 & 86.67 & \cellcolor{gray!30}53.33 & 71.43 & \cellcolor{gray!30}66.67 & \cellcolor{gray!30}50.00 & 91.43 & \cellcolor{gray!30}44.44 & 85.71 & \textbf{68.48} \\
Anthropic: Claude Opus 4.1 \citep{opus} & $-$ & 70.37 & \cellcolor{gray!30}60.00 & \cellcolor{gray!30}66.67 & 71.43 & \cellcolor{gray!30}58.33 & \cellcolor{gray!30}62.50 & 74.29 & 88.89 & 71.43 & \textbf{69.32} \\
Anthropic: Claude Sonnet 4 \citep{sonnet} & $-$ & \cellcolor{gray!30}62.96 & \cellcolor{gray!30}60.00 & \cellcolor{gray!30}33.33 & \cellcolor{gray!30}57.14 & \cellcolor{gray!30}41.67 & \cellcolor{gray!30}50.00 & 74.29 & \cellcolor{gray!30}11.11 & 78.57 & \textbf{52.12} \\
\midrule
\textbf{Category Average} & & \textbf{62.96} & \textbf{69.09} & \textbf{56.67} & \textbf{67.01} & \textbf{56.06} & \textbf{52.27} & \textbf{80.29} & \textbf{58.00} & \textbf{75.27} & \textbf{64.18} \\
\bottomrule
\end{tabular}
\label{tab:vit_performance2}
\end{table*}

Meta: Llama 3.2 90B Vision Instruct achieves the highest performance with only one pair-accuracy failure mode in Quantity and Count and no defined individually-measured failure modes, indicating stronger visual understanding and reasoning for similar images compared to other models. Conversely, xAI: Grok 4 performs the worst with only one category, Presence of Specific Features, above the threshold for failure modes.
Meta: Llama 4 Maverick and Meta: Llama 4 Scout are both from the same LLM family but contain key differences in architecture and structural setup. Maverick is attuned to high-performance generation and implementation with 17 billion active parameters for its 128 experts in its MoE (mixture-of-experts outlined in \citep{llama4}) architecture, totaling 400 billion parameters. This is larger than Scout’s input-focused MoE architecture with 17 billion active parameters and 16 experts, totaling 109 billion parameters. Mixture-of-experts utilizes gating networks, which essentially direct certain inputs to experts. Experts are smaller models meant for specific tasks that are part of the MLLM. The benefit of experts is that these smaller models can process the inputs without the entire MLLM having to be utilized, and this, in turn, would augment the MLLM's efficiency. Since the entire MLLM isn't being used, only some of its parameters are going to be active, and this is why, for example, Maverick only has 17 billion active parameters out of its 400 billion total parameters. On this note, Maverick's MoE architecture is represented as \texttt{17Bx128E}, whereas Scout's MoE architecture is represented as \texttt{17Bx16E}. However, both models perform relatively the same, with Maverick performing only slightly better.

\begin{table*}[htbp]
\caption{Individual Explicit Accuracies for Image Generation Models: Based on AMVICC (see Section \ref{igm_score_analysis}).}
\vskip 0.1in
\centering
\scriptsize
\setlength{\tabcolsep}{5pt}
\begin{tabular}{l|c|c|c|c|c|c|c|c|c|c|c}
\toprule
\textbf{Model} & \textbf{Params} & \faSync & \faCog & \faCompass & \textbf{A} & \faSort & \faMapPin & \faSearch & \faCamera & \faPalette & \textbf{Model} \\
& \textbf{Size (B)} & & & & & & & & & & \textbf{Average} \\
\midrule
OpenAI: DALL·E 3 \citep{dalle3} & $-$ & \cellcolor{gray!30}77.78 & 90.00 & \cellcolor{gray!30}66.67 & \cellcolor{gray!30}71.43 & \cellcolor{gray!30}66.67 & \cellcolor{gray!30}75.00 & \cellcolor{gray!30}75.71 & 83.33 & 89.29 & \textbf{77.32} \\
Google: Gemini 2.5 Flash Image \citep{gemini25flash} & $-$ & \cellcolor{cyan!25}94.44 & \cellcolor{cyan!25}96.67 & \cellcolor{cyan!25}96.67 & \cellcolor{gray!30}78.57 & \cellcolor{gray!30}75.00 & \cellcolor{cyan!25}90.63 & \cellcolor{cyan!25}85.71 & \cellcolor{cyan!25}100.00 & \cellcolor{cyan!25}96.43 & \textbf{90.46} \\
Stability AI: Stable Diffusion 3.5 Large \citep{stabled} & $8.1$ & \cellcolor{gray!30}55.56 & \cellcolor{gray!30}73.33 & \cellcolor{gray!30}56.67 & \cellcolor{gray!30}42.86 & \cellcolor{gray!30}50.00 & \cellcolor{gray!30}43.75 & \cellcolor{gray!30}67.14 & \cellcolor{gray!30}77.78 & \cellcolor{gray!30}78.57 & \textbf{60.63} \\
\midrule
\textbf{Category Average} & & \textbf{75.93} & \textbf{86.67} & \textbf{73.34} & \textbf{64.29} & \textbf{63.89} & \textbf{69.79} & \textbf{76.19} & \textbf{87.04} & \textbf{88.10} & \textbf{76.14} \\
\bottomrule
\end{tabular}
\label{tab:vit_performance3}
\end{table*}

\begin{table*}[htbp]
\caption{Pair Explicit Accuracies for Image Generation Models: Based on AMVICC (see Section \ref{igm_score_analysis}).}
\vskip 0.1in
\centering
\scriptsize
\setlength{\tabcolsep}{5pt}
\begin{tabular}{l|c|c|c|c|c|c|c|c|c|c|c}
\toprule
\textbf{Model} & \textbf{Params} & \faSync & \faCog & \faCompass & \textbf{A} & \faSort & \faMapPin & \faSearch & \faCamera & \faPalette & \textbf{Model} \\
& \textbf{Size (B)} & & & & & & & & & & \textbf{Average} \\
\midrule
OpenAI: DALL·E 3 \citep{dalle3} & $-$ & \cellcolor{gray!30}55.56 & 80.00 & \cellcolor{gray!30}40.00 & \cellcolor{gray!30}42.86 & \cellcolor{gray!30}50.00 & \cellcolor{gray!30}56.25 & \cellcolor{gray!30}57.14 & \cellcolor{gray!30}66.67 & 85.71 & \textbf{59.35} \\
Google: Gemini 2.5 Flash Image \citep{gemini25flash} & $-$ & \cellcolor{cyan!25}88.89 & \cellcolor{cyan!25}93.33 & \cellcolor{cyan!25}93.33 & \cellcolor{gray!30}57.14 & \cellcolor{gray!30}66.67 & \cellcolor{cyan!25}81.25 & \cellcolor{cyan!25}74.29 & \cellcolor{cyan!25}100.00 & \cellcolor{cyan!25}92.86 & \textbf{83.08} \\
Stability AI: Stable Diffusion 3.5 Large \citep{stabled} & $8.1$ & \cellcolor{gray!30}25.93 & \cellcolor{gray!30}46.67 & \cellcolor{gray!30}20.00 & \cellcolor{gray!30}14.29 & \cellcolor{gray!30}25.00 & \cellcolor{gray!30}12.50 & \cellcolor{gray!30}40.00 & \cellcolor{gray!30}66.67 & \cellcolor{gray!30}64.29 & \textbf{35.04} \\
\midrule
\textbf{Category Average} & & \textbf{56.79} & \textbf{73.33} & \textbf{51.11} & \textbf{38.10} & \textbf{47.22} & \textbf{50.00} & \textbf{57.14} & \textbf{77.78} & \textbf{80.95} & \textbf{59.16} \\
\bottomrule
\end{tabular}
\label{tab:vit_performance4}
\end{table*}

\subsection{IGM Score Analysis} \label{igm_score_analysis}
Across all three image generation models, two common failure modes appear in both individual explicit and pair explicit accuracies: Quantity and Count and Text (see Tables \ref{tab:vit_performance3} and \ref{tab:vit_performance4}). The majority of the models (2/3) also exhibit failure modes for both individual and pair explicit accuracies in the following categories: State and Condition, Orientation and Direction, Positional and Relational Context, and Presence of Specific Features.

Of the three models evaluated, Google: Gemini 2.5 Flash Image achieves the highest performance with only two failure modes across individual and pair explicit accuracies in Text and Quantity and Count. Inversely, Stability AI: Stable Diffusion 3.5 Large performs the worst, with all categories dropping below the standard for failure modes in both individual and pair explicit accuracies. In fact, Stable Diffusion 3.5 Large fails Positional and Relational Context with a pair explicit accuracy of 12.50\%, which is the lowest IGM accuracy recorded in our benchmark. Lastly, despite OpenAI: DALL·E 3 achieving moderate performance, its failure across 7 categories bolsters IGMs' shortcomings in generating images.

\subsection{Cross-Examination of IGMs and MLLMs}
Collectively, certain categories such as Quantity and Count constitute failures in IGMs and MLLMs, with both modalities performing notably poorly on them. Other common failure mode groupings include Positional and Relational Context, Orientation and Direction, and State and Condition. However, while both MLLMs and IGMs generally tend to fail in the same categories, there is one category that stands out as an exception to this trend: Text. MLLMs perform significantly better when processing textual contexts as majority of the models don't fail on Text. On the contrary, all 3 of the IGMs fail on Text for both their individual and pair explicit accuracies. This poor performance is compounded by the fact that pair accuracies for both MLLMs and IGMs are overall lower than all of the models' individual accuracies. This trend is apparent due to the requirement that both images or answers in a pair need to be accurate in order for them to be considered a correct pair.

Image generation models depict a larger disparity in the capabilities of each model: Stability AI: Stable Diffusion 3.5 Large is unable to follow elementary instructions in differentiating between the implicit and explicit prompts; meanwhile, Google: Gemini 2.5 Flash Image reliably adds components based on the explicit instructions. These results underscore the need for more intensive testing into the failure modes of MLLMs and IGMs in order to cross-reference influencing factors and improve visual intelligence and understanding across the field of machine learning.

\subsection{Ablation Studies}
To further explore the robustness and reliability of model behavior, we conduct a series of ablation studies designed to test sensitivity to prompt phrasing, model randomness, and architectural differences. These studies aim to isolate the factors that most influence success or failure across tasks.
\subsubsection{Linguistic Sensitivity}

In order to understand whether prompt wording and adaptation to questions directly affect the outcome and accuracies demonstrated from image generation models, we change the wording of 40 prompts and test them on OpenAI: DALL·E 3 to determine whether the accuracies would fall in the same range as the original tests. We utilize OpenAI's GPT-5 to improve prompt wording by adding context clues and disregarding the original prompt constraint of explicit prompts only having the new specific component in addition to their corresponding generic implicit prompts. We use a randomly generated interval of the prompts in order to ensure generalization of the sample to the population. However, based on the overall accuracy of the prompts, it is clear that adding more targeted language does not help improve model accuracy. Instead, it results in a noticeable, unexpected decrease in the pair accuracy for the Presence of Specific Features category in explicit prompts.

\begin{table}[htbp]
\vskip 0.062in
\captionof{table}{Linguistic Sensitivity Trials: Pair Implicit and Explicit Accuracies for Reworded Prompts. (C) denotes control/original wording; (W) denotes reworded prompts.}
\vskip 0.1in
\centering
\tiny
\setlength{\tabcolsep}{5pt}
\begin{tabular}{l|c|c|c|c|c|c}
\toprule
\textbf{Pair Implicit Types} & \faPalette & \faSearch & \faMapPin & \faSync & \faCog & \textbf{A} \\
\midrule
Pair Implicit (C) & 100.00 & 100.00 & 100.00 & 100.00 & 100.00 & 100.00 \\
Pair Implicit (W) & 100.00 & 100.00 & 100.00 & 100.00 & 100.00 & 100.00 \\
Pair Explicit (C) & 100.00 & 100.00 & 100.00 & 100.00 & 100.00 & 0.00 \\
Pair Explicit (W) & 100.00 & 75.00 & 100.00 & 100.00 & 100.00 & 0.00 \\
\bottomrule
\end{tabular}
\label{tab:linguistic_sensitivity}
\end{table}
\subsubsection{IGM Stochasticity}

To evaluate the significance of model stochasticity in IGMs, we test the 20 prompt pairs---the same 40 prompts that we reword for the Linguistic Sensitivity Trials---through 3 trials, generating 60 total implicit images and 60 total explicit images. We utilize OpenAI: DALL·E 3, the median-performing model between Google: Gemini 2.5 Flash Image and Stability AI: Stable Diffusion 3.5 Large, and run an identical experiment pipeline to the main experiment. Through the findings, we conclude that while prompts could individually vary in accuracy with certain prompts only scoring accurately on two of the three tests, individual variance does not drastically affect the overall accuracy of the test set in the sample. This highlights a negligible role of sampling variance in IGM failure modes and suggests that conceptual misunderstanding, rather than model stochasticity, accounts for the principal IGM accuracies.

\begin{table}[htbp]
\captionof{table}{IGM Stochasticity Trials: Individual and Pair Implicit and Explicit Accuracies for Three Separate Trials.}
\vskip 0.1in
\centering
\label{tab:vit_performance5}
\scriptsize
\begin{tabular}{l|c|c|c}
\toprule
\textbf{Tests} & \textbf{Test 1} & \textbf{Test 2} & \textbf{Test 3} \\
& & & \\
\midrule
Individual Implicit & 100.00 & 100.00 & 100.00 \\
Individual Explicit & 90.00 & 85.00 & 90.00 \\
Pair Implicit & 100.00 & 100.00 & 100.00 \\
Pair Explicit & 80.00 & 70.00 & 80.00 \\
\bottomrule
\end{tabular}
\end{table}

\section{Discussion}
Our findings indicate that IGMs generally exhibit equal or higher levels of failure compared to MLLMs. However, category-specific analysis reveals that performance still varies between the two, with each model type performing better in different category-specific tasks. Outliers on both ends of the spectrum include Meta: Llama 3.2 90B Vision Instruct and Google: Gemini 2.5 Flash Image, which achieve the best results, and xAI: Grok 4 and Stability AI: Stable Diffusion 3.5 Large, which demonstrate the worst performance of their respective modalities.

Each model exhibits fluctuations in performance compared to other models, alternating between producing stronger and weaker results. For instance, models of both modalities fail in Quantity and Count, but IGMs outperform MLLMs in Viewpoint and Perspective while MLLMs outperform IGMs in Text. However, if all of these models are trained on the same data structure and similar data (e.g., image-caption pairs in OpenAI: DALL·E 3), this could indicate that model size is not relevant to the elementary visual understandings of either VLMs or IGMs \citep{dalle3}.

Furthermore, Google: Gemini 2.5 Flash Image significantly outperforms Stability AI: Stable Diffusion 3.5 Large and OpenAI: DALL·E 3 in image realism and consistency. Consequently, a notable disparity in quality among image generation models emerges through our tests.

Nevertheless, as observed in human evaluation, all image generation models are often unable to leave out specific features in each category and are also unable to manipulate viewpoints to hide specific components as prompted, especially when “no” or “without” is included. This suggests that image generation models, despite the quality of their generated images, still struggle with elementary instruction-following for certain phrasing. Sometimes, components instructed to be partially hidden are fully shown, and components instructed to be fully hidden are still slightly seen, indicating that some generated images just barely fail to meet the entirety of their prompts' requirements; this reduces the overall accuracies of the image generation models. Even though Google: Gemini 2.5 Flash Image's capability far outperforms that of the other two IGMs, it still struggles with these same underlying issues that slightly diminish its accuracies. For instance, when all 3 models are asked to generate a keyboard for one of the prompts, the generated keyboard quality is drastically better and more realistic for Gemini 2.5 Flash Image compared to the other two models. However, all three models fail to follow the implied instruction when prompted to create an image in contexts of greater difficulty, where it isn't systematically stated how to achieve the image. This limits their ability to accomplish the prompt's direct requirement of having or not having a specific element in the generated image. For example, one of the explicit prompts\footnote{An explicit prompt specifies the content that must be included in the image, whereas explicit instruction specifies how that content should be achieved or generated.} instructs the IGMs to generate a computer keyboard with the Z key hidden. In the prompt, it is not expressly stated that the IGMs have to orient the image angle in a way where the Z key is hidden; the models have to understand the implied instruction in order to satisfy the prompt.

Some image generation models also indicate struggles with understanding contextual cues and alignment pertaining to natural human thought. For instance, if asked to produce a stripe down the middle of a car, Stability AI: Stable Diffusion 3.5 Large would produce a stripe across the horizontal middle of the car, while OpenAI: DALL·E 3 and Google: Gemini 2.5 Flash Image would produce a stripe across the vertical middle of the car, as many humans would naturally think.

Interestingly, the architectures of the best and worst-performing models of different modalities offer key insights and introduce new questions about the relevance of various architectures in model performance for elementary visual understanding and depiction. For example, Meta: Llama 3.2 90B Vision Instruct---whose architecture consists of a two-stage vision encoder added on to a frozen LLM---easily outperforms OpenAI's GPT-4o across all but two categories: Text and Color and Appearance, despite typically not outperforming the more popular VLMs such as GPT-4o on complex tasks. In terms of IGMs, Google: Gemini 2.5 Flash Image, a sparse mixture-of-experts (MoE) transformer, outperforms OpenAI: DALL·E 3 even though they are both built with a natively multimodal architecture and trained on similarly structured pairs of image and text data.

These inconsistencies could create systems-level deployment challenges due to a lack of accuracy in elementary reasoning, leading to long-term oversights in basic tasks and essentially risking efficiency and scalability. It is necessary to perform more in-depth testing to uncover the basis for why image generation models and multimodal LLMs seem to fail and succeed in differing categories. We hope that our work provides the foundational data to understand where current models fail and succeed.

\section{Related Works}
\subsection{Failure Modes in Image Generation Models}
Text-to-image generation models such as OpenAI: DALL·E 3 and Stability AI: Stable Diffusion 3.5 Large have made rapid progress in image quality but continue to face challenges in commonsense reasoning, fairness, and scene composition. Recent evaluations have shown systematic biases and reasoning failures in these models, raising questions about their true semantic understanding. Commonsense-T2I Challenge shows major failures in reasoning: DALL·E 3 attains an accuracy of only approximately 49\% \citep{commonsense}. A biased survey identifies a lack of evaluation frameworks and coverage of non-binary identities \citep{bias}. Similarly, a diffusion model survey highlights specific weaknesses like generating multiple objects and rare concepts; proposed layout and attention improvements seek to improve the model \citep{diffusion}. Although this work identifies critical weaknesses in generative performance, it remains unclear whether these weaknesses are shared with interpretive failures in vision language models or whether they have been directly compared to correlating tasks within varied-architecture MLLMs.

\subsection{Visual Reasoning Challenges in Visual Language Models}
Visual language models (VLMs) like OpenAI: GPT-4o and Google: Gemini 2.5 Pro have become central to visual reasoning tasks, yet they often falter on simple image-based questions. Efforts to improve VLMs are centered around better pretraining, alignment, and hallucination reduction using approaches like VILA, CogVLM2, and SIMA. VILA shows improved in-context learning and world knowledge from interleaved pretraining \citep{pretraining}. SIMA reduces hallucinations and boosts VQA benchmark accuracy via visual critic metrics \citep{alignment}. CogVLM2 achieves SOTA across multiple visual benchmarks with an efficient architecture \citep{cogvlm2}. Despite these advances, prior work focuses solely on improving VLMs without evaluating whether these errors also emerge during generative tasks. Current existing studies don't test model performance on aligned image/question pairs.

\section{Conclusion}
In this work, we introduce a novel benchmark, AMVICC (Assessment of Modality-Specific Visual Intelligence Comprehension and Creation), to evaluate the cross-modal failure modes of multimodal large language models and image generation models in order to gain insight into the commonalities and distinctions. We conclude that not only do IGMs and MLLMs share certain common failure modes and differ on others, but they also diverge within specific modalities to create model-specific failure modes that can be attributed to a wide range of factors. Future work can expand the MMVP or AMVICC benchmarks to increase the range of visual understanding categories evaluated or to improve visual understanding on specific models to augment accuracy for specific categories. Further extensions of this paper can replicate tests to prove accuracy on a larger scale with more resources.

\section{Limitations}
\subsection{Methodology Limitations}
The primary limitation present within this methodology is the conversion from the MMVP questions to specific prompts that cover the same visual elements as the questions. As the prompts are written by four authors of our research team, albeit following a strict linguistic structure, there is inherent prompt design bias. This hinders our ability to definitively state that the translation of categories and tasks tested can be completely translated to image generation models. However, the procedure that we utilize to define the creation of the prompts, as outlined in Section \ref{prompting_procedure}, ensures that each prompt follows the same structure and inherits the same information from each corresponding question to facilitate rigorous alignment.

Furthermore, each prompt and question could fall under multiple categories. Nonetheless, to allow for predominantly accurate findings, we assign each prompt and question to only one category. Consequently, while performance on the prompts and questions could also influence the accuracy of other categories that they could fall under, it is not incorporated into the final numbers. Additionally, each prompt is double-checked by multiple human prompt writers to optimize categorization and mitigate this issue.

Another limitation includes the unbalanced model usage of IGMs compared to MLLMs. Due to the lack of availability of image generation models through API keys and time constraints, we are unable to test as many IGMs as MLLMs. This imbalance means that our accuracy averages for our IGMs could potentially be less representative of the overall failure modes of all IGMs, when compared to the representation offered by our MLLM accuracy averages.

\subsection{Evaluation Limitations}
Due to our MLLMs having been proven to have visual reasoning deficiencies, we choose to use human evaluators to determine the accuracy of the outputs produced by image generation. Despite the rubrics outlined in Tables \ref{tab:rubric_igm} and \ref{tab:rubric_vlm} to reduce subjectivity of human evaluators, there is still a chance of human subjectivity bias in the results. However, the specificity of the rubric limited the ability of the empirical data to represent the confounding factors of the data, such as the situational factors generated around the specific criteria (e.g., the Z key in a keyboard compared to an inaccurate depiction of a keyboard is still incorrect). These, due to computational power and human resources, limit the extent to which the failure modes can be understood from the data.

Furthermore, OpenAI's GPT-4 is utilized as a grader for the MLLMs. This could skew the results due to a lack of a human counterpart in evaluations, and because there are answer choices present, the AI grader would often be generalizing any potential responses from an MLLM into either (a) or (b) as an answer.

Another evaluation limitation encompasses the lack of a human performance control group for image generation performance due to the technological nature of the task that we are testing on IGMs. This requires us to understand the competencies and capabilities of models through relational comparison between models.

The last evaluation limitation arises from the closed-source nature of many of the models. We are unable to look at the internal elements of the model and must only rely on surface-level documentation provided by commercial companies (e.g., OpenAI's DALL·E 3).

\section*{Impact Statement}

This paper presents work whose goal is to advance the field of Machine
Learning. There are many potential societal consequences of our work, none
which we feel must be specifically highlighted here.


\bibliography{example_paper}
\bibliographystyle{icml2026}

\clearpage
\appendix
\section {Extended Methodology and Supplemental Results} \label{appendix_a}
\subsection {Prompt Set for Image Generation}
Below is a link to the prompts used to guide the image generation process. The prompts are adapted directly from the MMVP benchmark to ensure consistency across tasks: \href{https://github.com/AahanaB24/AMVICC/blob/main/scripts/AMVICC.csv}{AMVICC Prompt Set}
\subsubsection{Categories (Defined):}
\begin{enumerate}
\item Orientation and Direction (od): The model’s ability to accurately detect the position, alignment, facing direction, or angles of objects in the image.
\item Presence of Specific Features (pf): The ability of a model to identify if specific visual characteristics, objects, or fine-grained attributes are explicitly present in an image.
\item State and Condition (sc): This refers to the model’s ability to be able to recognize the current status, phase, or physical condition of an object, entity, or scene that is being depicted in an image.
\item Quantity and Count (qc): The model’s ability to identify the number of objects, people, or elements in an image, including the tasks that involve counting, estimating quantities, or comparing amounts.
\item Positional and Relational Context (pr): This refers to a model’s ability to be able to understand the spatial relationships and relative positions between objects or entities within an image.
\item Color and Appearance (ca): This refers to the ability of the model to perceive, recognize, and reason about colors, visual patterns, and image-level characteristics like tone, brightness, and artistic style.
\item Structural and Physical Characteristics (sh): The model’s ability to perceive and reason about the shape, material, construction, and physical properties of objects or elements within an image.
\item Text (tx): The ability of a model to detect, recognize, and interpret written language (printed, handwritten, or stylized text) that appears within an image and to reason about its content, meaning, and context.
\item Viewpoint and Perspective (vp): This refers to the ability of a model to be able to recognize and reason about the camera's or observer’s perspective and angle relative to the objects or scene in an image, affecting how elements are visually presented.
\end{enumerate}
Each task (image interpretation or image generation) is analyzed independently and comparatively across these 9 dimensions to identify common and divergent failure modes.
\subsection {Code Base}
All code used in this study for model evaluation, result collection, and visualization is available at: \href{https://github.com/AahanaB24/AMVICC}{AMVICC GitHub Repository}
\subsection {Experiments (Further Outlined)}
This section outlines how we apply our methods to test the failure mode alignment between vision language models and image generation models, specifying the experimental conditions, controls, and design decisions that underpin our analysis.

\textit{Overview and Hypotheses:}
We test the core hypothesis: Do the failure modes of VLMs in visual reasoning correlate with the failure modes of IGMs when tasked with generating images that express those same visual concepts?

\textit{This hypothesis rests on two premises:}
If VLMs fail to understand a visual concept (e.g., object orientation), then IGMs may also fail to generate that concept reliably.
Alternatively, divergence in failure patterns would suggest modality-specific weaknesses, pointing to differences in model architecture or training objectives.

\textit{Experimental Variations and Comparative Design:}
To probe our hypothesis and ensure robustness, we introduce several comparative and diagnostic experiments:
Cross-Modality Comparison:
VLM Task: Answer MMVP questions based on real and generated images.
IGM Task: Generate images based on prompts derived from MMVP questions.
Explicit vs. Implicit Prompting:
We vary prompt specificity to test if IGMs struggle more with indirect language.
This also enables assessment of whether image failures propagate into VLM misinterpretation when fed generated content.

\textit{Ablation: Prompt Rewording:}
For failure-prone prompts, we create reworded versions to test whether small linguistic changes improve image generation accuracy or alter failure types.

\textit{Ablation: Repetition Analysis (Randomness Test):}
For 30 selected prompts, OpenAI: DALL·E 3 is queried 5 times each.
We analyze image generation consistency and its impact on downstream VLM accuracy.

\textit{Ablation: Architecture/Scale Variation:}
We include related models with different parameter sizes (e.g., Llama 4 Maverick vs. Scout) to evaluate the impact of architecture vs. scale.
All of these comparisons allow us to not only isolate when the models fail, but they also allow us to recognize why the models fail, whether it is due to conceptual, linguistic, architectural, or visual representation limitations.

\textit{Data Summary:}
Below is a description of the experiment flow:
VLM Baseline: Each of the 300 MMVP (image, question) pairs is evaluated across 11 VLMs.
IGM Prompting: 600 text prompts (explicit + implicit) are derived from MMVP questions and serve as inputs to 3 IGMs.

\textit{Generated Image Evaluation:}
Human annotators judge whether the generated images accurately capture the core visual concepts.
Generated images are fed back into VLMs to answer the original questions.

\textit{Cross-analysis:}
Accuracies and failure mode types are tracked and compared across tasks, models, categories, and prompt types.
\subsection{VLM Image Analysis Responses}

When prompted to analyze the images from the MMVP dataset, the VLMs produce outputs containing their answers and explanations. These responses are used to calculate individual and pair accuracies for VLM evaluation. One of the responses includes:

\begin{small}
\begin{verbatim}
{
  "question_id": 26,
  "category": "ca",
  "question": "What color is the chicken's body?
               (a) Black (b) Red",
  "correct_answer": "(b)",
  "model_response": "The chicken's body is a
                     reddish-brown color, which
                     is typical for certain
                     breeds like Rhode Island
                     Reds. So, the correct
                     answer is: (b) Red",
  "is_correct": true,
  "gpt_grade": "yes"
}
\end{verbatim}
\end{small}

\subsection {Rubric for Image Generation \& Vision Language Model Evaluation}
Below are the rubrics used to assess whether a generated image successfully follows a prompt and whether a VLM correctly answers a question:

\begin{table}[h]
\captionof{table}{Rubric for Image Generation Evaluation}
\vskip 0.1in
\centering
\includegraphics[width=\linewidth]{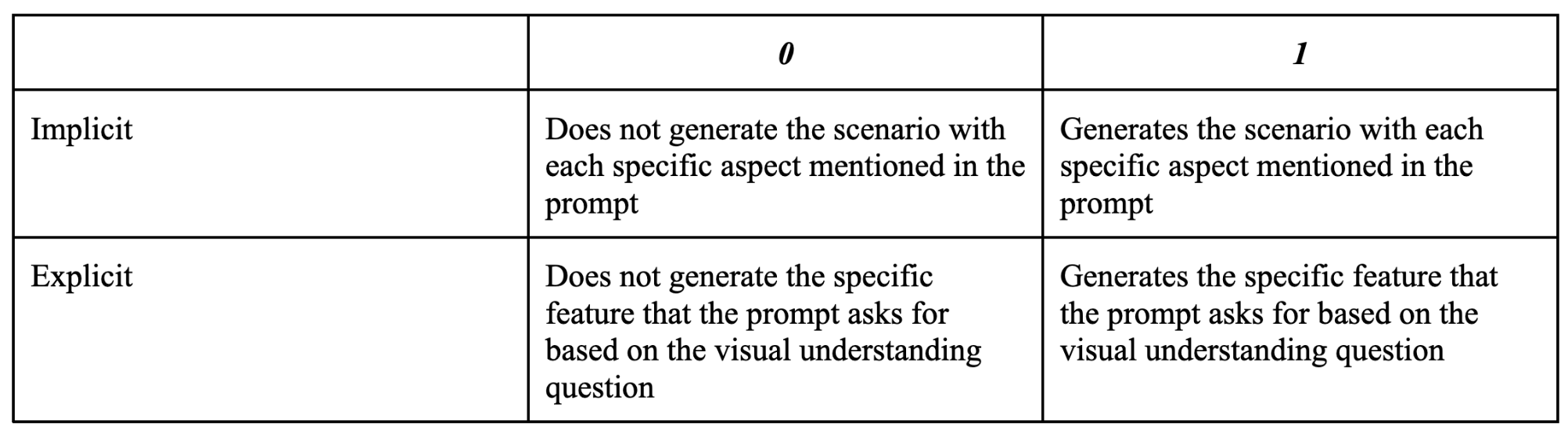}
\label{tab:rubric_igm}
\end{table}

\begin{table}[h]
\captionof{table}{Rubric for Vision Language Model Evaluation}
\vskip 0.1in
\centering
\includegraphics[width=\linewidth]{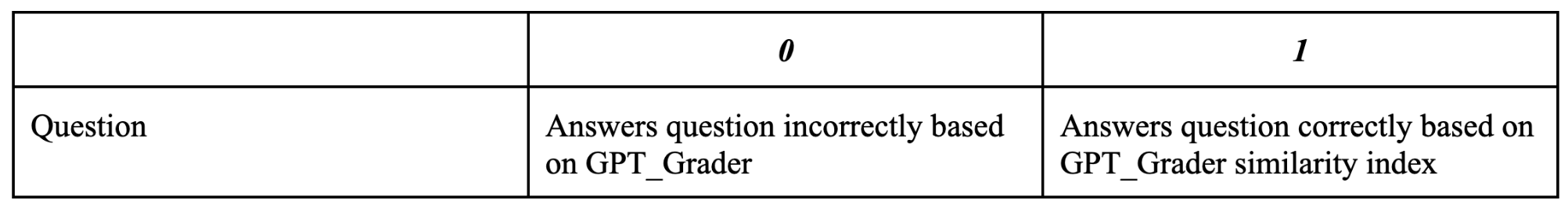}
\label{tab:rubric_vlm}
\end{table}

\subsection {Generated Image Results \& Extra Results}
\begin{figure*}[htbp]
\centering

\colorbox{gray!10}{%
\begin{minipage}{0.98\textwidth}
\vspace{0.3cm}
\begin{subfigure}[t]{0.18\textwidth}
\centering
\parbox[t][1.8cm]{\textwidth}{\centering\footnotesize\textbf{Can you see the key "Z" in the image?}}
\vspace{0.6cm}
\includegraphics[width=\textwidth, height=2.5cm]{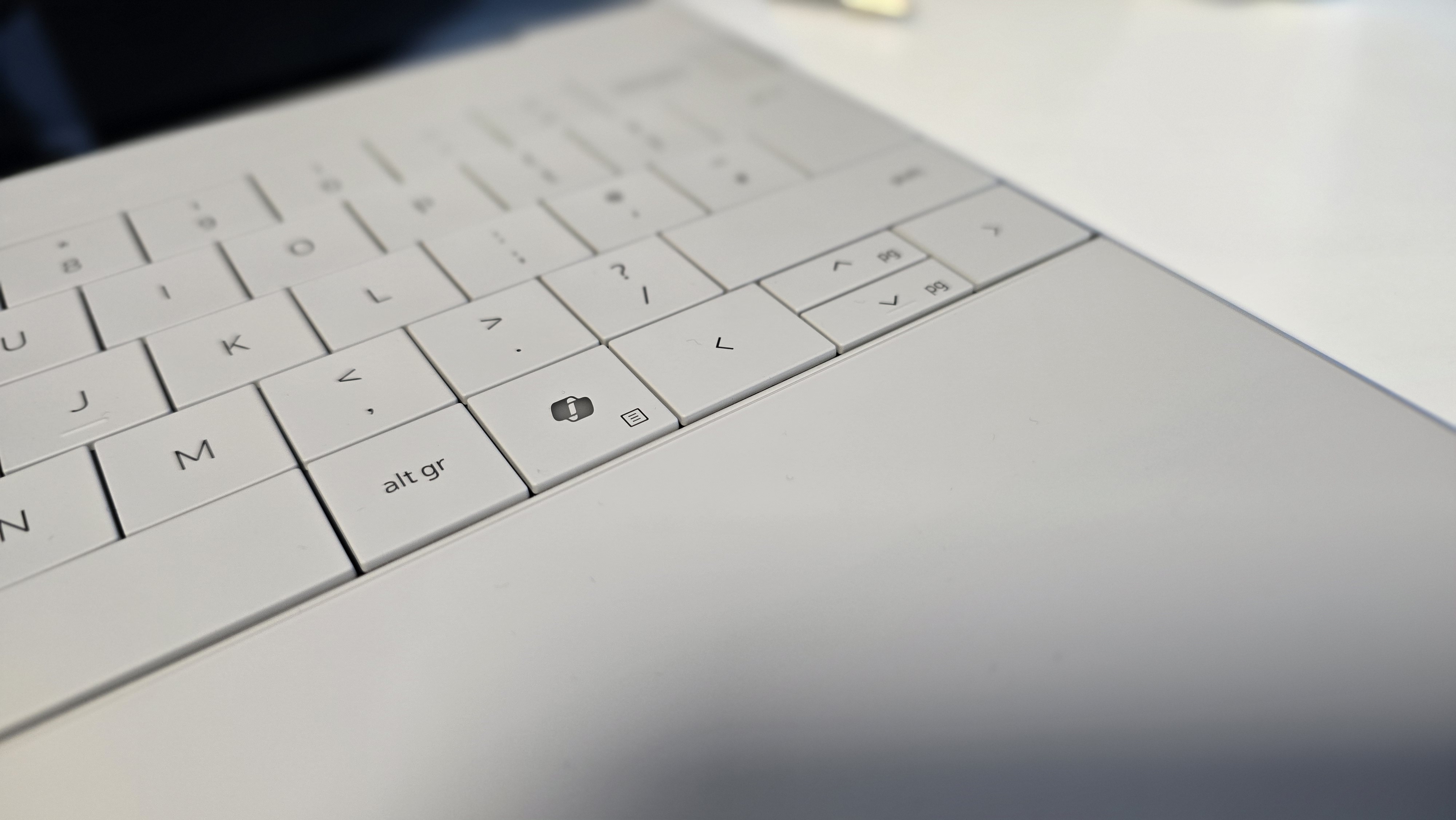}
\vspace{0.3cm}
\parbox[t][0.8cm]{\textwidth}{\centering\footnotesize (a) Yes \\ \colorbox{cyan!25}{(b) No}}
\vspace{0.1cm}
\parbox{\textwidth}{\centering\scriptsize\textit{Prompt: Generate a computer keyboard photographed at an angle where the Z key is hidden}}
\end{subfigure}
\hfill
\begin{subfigure}[t]{0.18\textwidth}
\centering
\parbox[t][1.8cm]{\textwidth}{\centering\footnotesize\textbf{Is the shark's belly visible in this image?}}
\vspace{0.6cm}
\includegraphics[width=\textwidth, height=2.5cm]{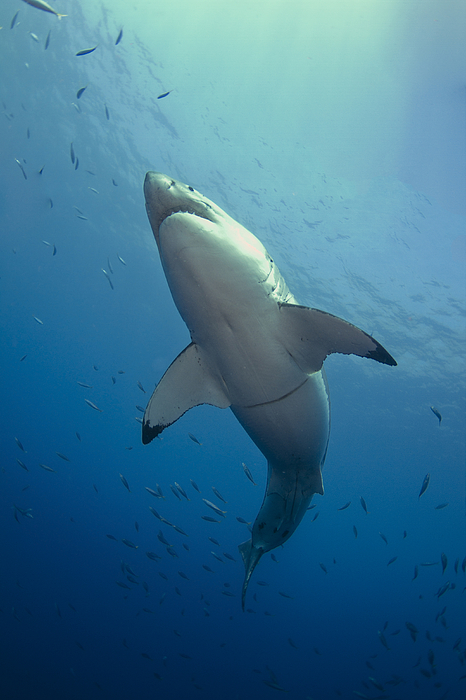}
\vspace{0.3cm}
\parbox[t][0.8cm]{\textwidth}{\centering\footnotesize \colorbox{cyan!25}{(a) Yes} \\ (b) No}
\vspace{0.1cm}
\parbox{\textwidth}{\centering\scriptsize\textit{Prompt: Generate a shark photographed from below showing its belly}}
\end{subfigure}
\hfill
\begin{subfigure}[t]{0.18\textwidth}
\centering
\parbox[t][1.8cm]{\textwidth}{\centering\footnotesize\textbf{Does the elephant have long or short tusks?}}
\vspace{0.6cm}
\includegraphics[width=\textwidth, height=2.5cm]{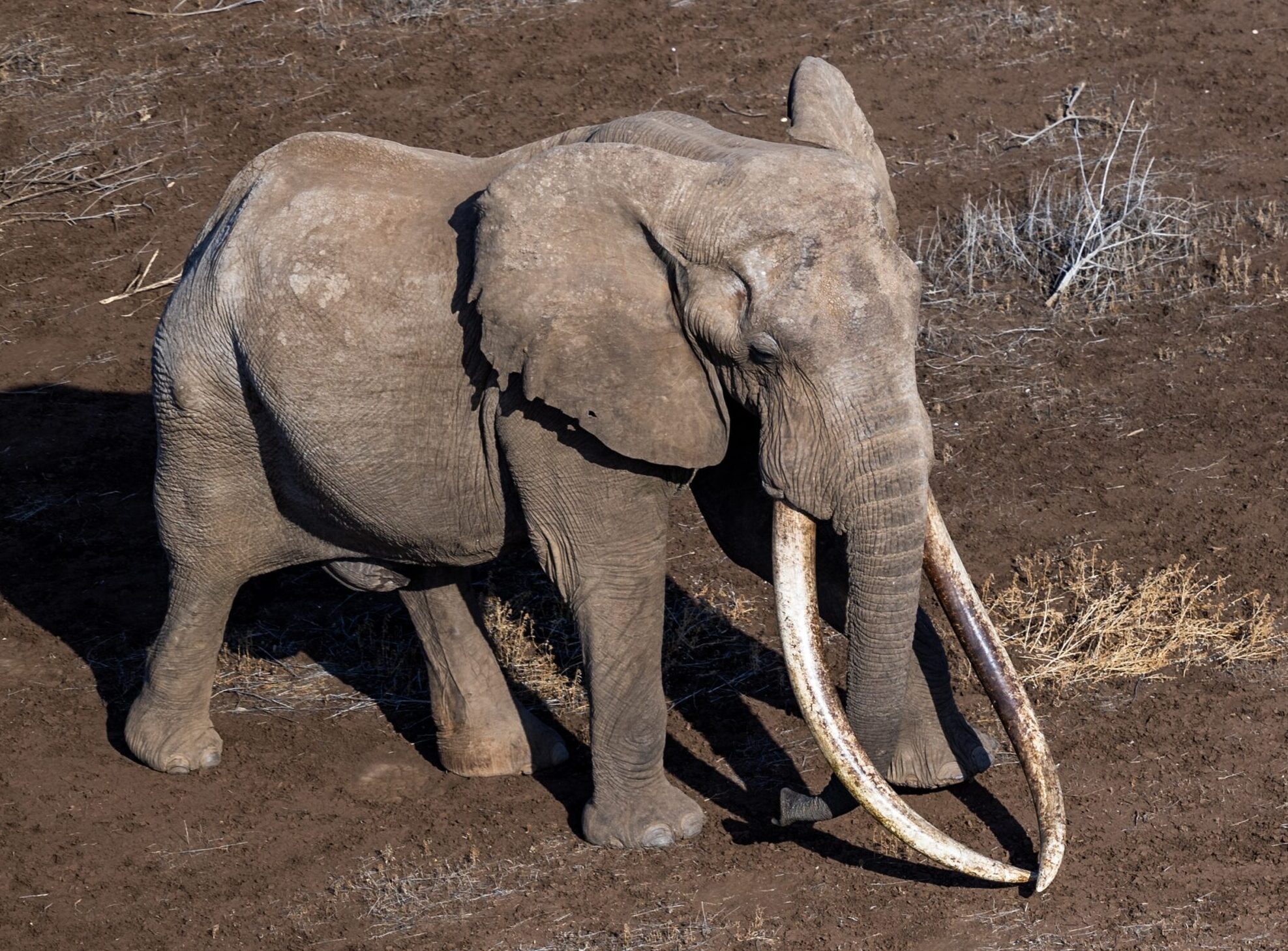}
\vspace{0.3cm}
\parbox[t][0.8cm]{\textwidth}{\centering\footnotesize \colorbox{cyan!25}{(a) Long} \\ (b) Short}
\vspace{0.1cm}
\parbox{\textwidth}{\centering\scriptsize\textit{Prompt: Generate an elephant with prominently long curved tusks}}
\end{subfigure}
\hfill
\begin{subfigure}[t]{0.18\textwidth}
\centering
\parbox[t][1.8cm]{\textwidth}{\centering\footnotesize\textbf{Is there a lemon inside the drink in the cup, or are all the lemons outside the drink?}}
\vspace{0.6cm}
\includegraphics[width=\textwidth, height=2.5cm]{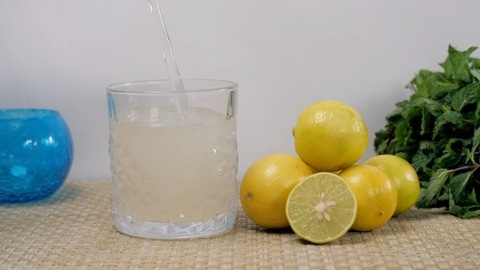}
\vspace{0.3cm}
\parbox[t][0.8cm]{\textwidth}{\centering\footnotesize (a) There is one inside \\ \colorbox{cyan!25}{(b) All are outside}}
\vspace{0.1cm}
\parbox{\textwidth}{\centering\scriptsize\textit{Prompt: Generate water in a glass with mint leaves and lemon slices, with all the lemons outside of the glass}}
\end{subfigure}
\hfill
\begin{subfigure}[t]{0.18\textwidth}
\centering
\parbox[t][1.8cm]{\textwidth}{\centering\footnotesize\textbf{Are there any words displayed on the vehicle's lightbar?}}
\vspace{0.6cm}
\includegraphics[width=\textwidth, height=2.5cm]{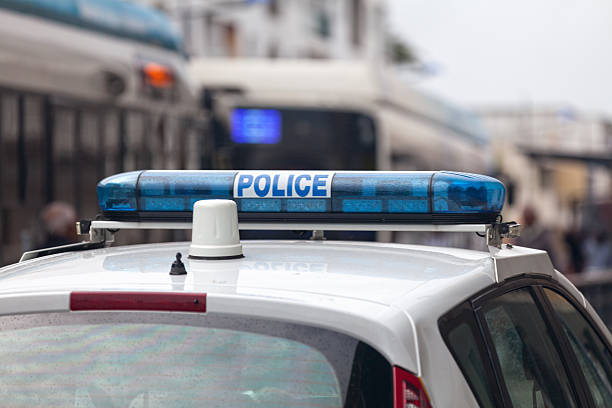}
\vspace{0.3cm}
\parbox[t][0.8cm]{\textwidth}{\centering\footnotesize \colorbox{cyan!25}{(a) Yes} \\ (b) No}
\vspace{0.1cm}
\parbox{\textwidth}{\centering\scriptsize\textit{Prompt: Generate a police car with "POLICE" text clearly visible on the lightbar}}
\end{subfigure}

\vspace{0.8cm}

\begin{adjustbox}{width=0.95\textwidth,center}
\small
\begin{tabular}{l|c c|c c|c c|c c|c c}
\toprule
& \multicolumn{2}{c|}{Question 1} & \multicolumn{2}{c|}{Question 2} & \multicolumn{2}{c|}{Question 3} & \multicolumn{2}{c|}{Question 4} & \multicolumn{2}{c}{Question 5} \\
\cmidrule(lr){2-3} \cmidrule(lr){4-5} \cmidrule(lr){6-7} \cmidrule(lr){8-9} \cmidrule(lr){10-11}
\textbf{Model} & \textbf{Ans.} & \textbf{Result} & \textbf{Ans.} & \textbf{Result} & \textbf{Ans.} & \textbf{Result} & \textbf{Ans.} & \textbf{Result} & \textbf{Ans.} & \textbf{Result} \\
\midrule
OpenAI: DALL·E 3 & a & \xmark & a & \cmark & a & \cmark & b & \cmark & b & \xmark \\
Google: Gemini 2.5 Flash Image & a & \xmark & a & \cmark & a & \cmark & b & \cmark & a & \cmark \\
Stability AI: Stable Diffusion 3.5 Large & b & \cmark & a & \cmark & b & \xmark & a & \xmark & b & \xmark \\
\bottomrule
\end{tabular}
\end{adjustbox}

\vspace{0.3cm}
\end{minipage}%
}

\vskip 0.15in
\caption{Examples of specific IGMs' abilities to generate an image based on an explicit prompt. We handpick 5 out of the 300 questions in the MMVP dataset to delineate disparities between the models. It is apparent that Google: Gemini 2.5 Flash Image is the most accurate, followed by OpenAI: DALL·E 3, and Stability AI: Stable Diffusion 3.5 Large, in that order. An important thing to note is that the IGMs don't directly state Yes or No or any of the answer choices, for that matter. However, based on the models' image generation, we can associate certain answer choices with the models. A \cmark~indicates that the model generated an image in accordance with the given prompt, whereas an \xmark~indicates the opposite.}
\label{fig:vqa_comparison}
\end{figure*}

Figure \ref{fig:vqa_comparison} below outlines 5 images generated from 5 of the prompts\footnote{All of the prompts are available in scripts/AMVICC.csv within the AMVICC GitHub repository.}. These outputs are used as part of the image analysis phase to assess whether image generation models can accurately depict the components in the prompts. Below is a link to the GitHub repository that houses the individual and pair implicit and explicit accuracies, as produced by the image generation evaluation code: \href{https://github.com/AahanaB24/AMVICC}{AMVICC GitHub Repository}

\vspace{0.25cm}

\noindent\textbf{More results available on Zenodo: \href{https://zenodo.org/records/17646068}{AMVICC Results \& Evaluations}}

\clearpage

\label{sec:appendix}

\end{document}